\documentclass[dvipdfmx]{interact}
\usepackage{amsmath}
\usepackage{amsfonts}
\usepackage{diagbox}
\usepackage{comment}
\usepackage{epstopdf}
\usepackage[caption=false]{subfig}

\usepackage[numbers,sort&compress]{natbib}
\bibpunct[, ]{[}{]}{,}{n}{,}{,}
\renewcommand\bibfont{\fontsize{10}{12}\selectfont}
\makeatletter
\def\NAT@def@citea{\def\@citea{\NAT@separator}}
\makeatother

\theoremstyle{plain}

\theoremstyle{definition}

\theoremstyle{remark}

\usepackage{color}
\usepackage{algorithm}
\usepackage{algpseudocode}
\usepackage{enumerate}
\usepackage{url}

\newcommand{\Add}[1]{\textcolor{black}{#1}}	
\newcommand{\Erase}[1]{\if0{#1}\fi}	

\begin{document}

\interfootnotelinepenalty=1000

\articletype{FULL PAPER}

\title{A Framework of Explanation Generation toward Reliable Autonomous Robots}

\author{
\name{Tatsuya Sakai\textsuperscript{a}\thanks{CORRESPONDING AUTHOR: Takayuki Nagai. Email: nagai@sys.es.osaka-u.ac.jp}, Kazuki Miyazawa\textsuperscript{a}, Takato Horii\textsuperscript{a} and Takayuki Nagai\textsuperscript{a,b}}
\affil{\textsuperscript{a}Graduate School of Engineering Science, Osaka University, Osaka, Japan;\\
\textsuperscript{b}Artificial Intelligence Exploration Research Center, The University of Electro-Communications, Tokyo, Japan}
}

\maketitle

\begin{abstract}
To realize autonomous collaborative robots, it is important to increase the trust that users have in them. Toward this goal, this paper proposes an algorithm which endows an autonomous agent with the ability to explain the transition from the current state to the target state in a Markov decision process (MDP). According to cognitive science, to generate an explanation that is acceptable to humans, it is important to present the minimum information necessary to sufficiently understand an event. To meet this requirement, this study proposes a framework for identifying important elements in the decision-making process using a prediction model for the world and generating explanations based on these elements. To verify the ability of the proposed method to generate explanations, we conducted an experiment using a grid environment. It was inferred from the result of a simulation experiment that the explanation generated using the proposed method was composed of the minimum elements important for understanding the transition from the current state to the target state. Furthermore, subject experiments showed that the generated explanation was a good summary of the process of state transition, and that a high evaluation was obtained for the explanation of the reason for an action.
\end{abstract}


\section{Introduction}
The machine learning technology has varied applications, such as medical diagnosis and image restoration \cite{medical,image_fix}, and its ability to execute tasks is remarkable. Naturally, it is applied in the field of robotics, and the day when autonomous robots that make sophisticated decisions will permeate our daily lives may not be far away. However, autonomous robots will probably not gain widespread acceptance solely through technological developments aimed at improving the accuracy of task achievement. It is necessary to increase the sense of trust in robot decision-making to popularize autonomous robots as our partners in daily life, rather than as systems to be wielded as tools. In other words, autonomous robots must be able to accurately present the information that a user wants to know about their behavior. 

Let us imagine a scenario in which a user asks a domestic robot to wash a shirt, and the robot removes the unwashed shirt
\Add{that the user 
put in the washing machine.}
When asked why, the robot replies, ``Because the shirt was in the washing machine.'' Although this explanation provides a basis for the robot's decision, it may not be the answer expected by the user. In this situation, it would be effective to present information on future actions, such as ``soaking the shirt in warm water first will make it easier to remove the dirt,'' by way of explanation.

\begin{figure}[t]
  \centering
    \includegraphics[width=0.9\linewidth]{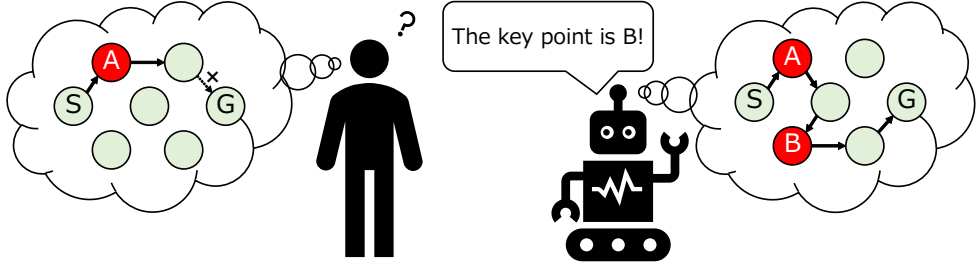}
    \caption[Research outline] {Research outline. The agent estimates the transition of the observed state and identifies the important elements. By supplementing the gap in the questioner's knowledge, the agent generates an explanation that satisfies likeliness and loveliness.} 
    \label{fig:intro}
\end{figure}

As can be seen from this example, the explainability required of autonomous robots is different in nature from the problems being addressed in the area of explainable artificial intelligence (XAI). 
XAI is primarily used by knowledgeable users to debug their systems, and is only useful when it is assumed that the user can derive the desired decision-making results from the input features \cite{XAIsurvey2018}. However, for humans to accept the independent existence of autonomous robots, it is sometimes necessary to explain the flow from the input features to the desired decision-making result. 
In the previous example, the user must understand that the shirt reaches the state of being washed according to the flow of ``inside the washing machine'' $\rightarrow$ ``outside the washing machine'' $\rightarrow$ ``soaking in a bucket'' $\rightarrow$ ``washing in the washing machine'' $\rightarrow$ ``washed''. 
The robots that can present such explanations are called explainable autonomous robots (XAR) \Add{\cite{surveyXAR}}. 
\Erase{
The problem of autonomous robots generating explanations can be framed as follows.
\begin{description}
    \item [Problem setting: Explanation generation by autonomous robot]\mbox{}\\
    Assume that a Markov decision process (MDP) is the framework for action determination used by the autonomous robot and user, and let the current state and the target state be shared between the robot and the user (Figure \ref{fig:intro}). The task of providing the reason for selecting an action in the current state is reduced to the problem of giving the necessary information for the user to estimate the transition from the current state to the target state based on the robot's action until the target state is achieved and the transition of states accompanying the actions.
\end{description}
}

In this paper, we propose an explanation generation framework for robots and other general autonomous agents. 
\Add{
The concept of explanation assumed in this study is shown in Figure \ref{fig:intro}. 
}
First, let the agent learn the prediction model of the world and the transition from the current state to the target state be possible to estimate without actual action. Using the world prediction model, the approximate causal effect of each action on the attainment of the goal is obtained, and the elements with high causal effect are extracted as important elements. 
Furthermore,
\Add{the action sequence assumed by the user is provided to the explanation generation model as a query, and}
the elements that are not understood by the user are 
\Erase{estimated and}
presented as explanations\footnote{\Add{Estimating the action sequence and the decision-making space held by the user is important for the generation of explanations. See \cite{surveyXAR} for more information on these research topics, which are beyond the scope of this study.}}.
We verified the usefulness of the proposed method through simulation using a grid maze environment. The effectiveness of the proposed method was evaluated by verification using a virtual agent given the explanation, and by conducting an experiment in which explanations are actually provided to subjects. 

\Add{
The rest of this paper is organized as follows. 
First, the research problem is formulated in the next section. 
Related works are reviewed, and this study is positioned with respect to them in Section 3.
}
In Section 4, the details of the proposed method are presented. 
Simulation and subject experiments are described in Sections 5 and 6, respectively. In Section 7, a discussion based on these results is presented. Section 8 presents a summary of the paper.

\Add{
\section{Problem Formulation}
In this study, we adopted the MDP framework for the agent's decision-making process to examine the explanations accepted by humans. In the MDP framework, the agent's decision-making is modeled using the following elements \cite{Madumal2020DistalEF}.
\begin{itemize}
    \item $\mathcal{A}$ : Set of actions that the agent can select
    \item $\mathcal{S}$ : Set of states
    \item $\mathcal{R}:\mathcal{S}\times \mathcal{A}\to \mathbb{R}$ : Reward function
    \item $\mathcal{T}=\{P(s'\mid a,s)\}$ : The probability of transition to the state $s'\in \mathcal{S}$ when performing action $a\in \mathcal{A}$ in a state  $s\in \mathcal{S}$
\end{itemize}
Here, the subset of the state-action pairs $(s,a)$ on an MDP is defined as explanation $E$. The purpose is to explain the process of reaching the target state $s_{target}$ in a form that is acceptable to humans. 
To do so, 
it is better to present the minimum information necessary to understand the path to the target state in the graph. 
That is, the set $E$ of state-action pairs that satisfy the following equation is obtained as follows:
\begin{equation}
\label{definition}
\mbox{$\min |E| ,\ subject\ to\  P(Reach(s_{target})\mid E)>\alpha$}
\end{equation}
Here, $Reach(s_{target})$ is the event of reaching the target state $s_{target}$, and $\alpha\in[0,1)$ is the level of comprehension required by the recipient of the explanation. Furthermore, $|E|$ denotes the number of elements in the set $E$. 
To solve this optimization problem, it is necessary to design a clear method to determine the questioner's level of understanding and the probability of reaching the target state.
However, a clear definition is exceedingly difficult. Therefore, in this study, we considered a heuristic solution in which $\rm(\hspace{.10em}i\hspace{.10em})$~the important elements for reaching the target state are identified, and $\rm(\hspace{.10em}ii\hspace{.10em})$~the elements that the user does not understand are presented. Each can be examined using an MDP as follows:
\begin{description}
    \item [$\rm(\hspace{.10em}i\hspace{.10em})$ Important elements for reaching the target state]\mbox{}\\
    An important element for reaching the target state is ``an element that is useful for imagining the process of reaching the target state and is absolutely necessary when reaching the target state.'' Therefore, ``a state-action pair in which it is difficult to find an alternative route to the target state on the MDP'' can be defined as an important element.
    \item [$\rm(\hspace{.10em}ii\hspace{.10em})$ Elements that the user does not understand]\mbox{}\\
    If the user understands all the transitions to the target state in the MDP, they can understand the process through which each action leads toward the target. Based on this, the elements that the user does not understand are ``elements for which the user cannot correctly estimate the transition in the MDP.''
\end{description}
}

\section{Related Works}
\subsection{Index to measure the goodness of explanation}
In this paper, we aim to explain the progress from the current state to the target state of the agent in a form that is easily accepted by humans. For that purpose, it is necessary to set an index to measure the goodness of explanation, which falls within the ambit of cognitive science. In cognitive science, two concepts indicate the goodness of a description: likeliness and loveliness \Add{\cite{explanatory,lipton}}. 
Likeliness is an indicator of the goodness of explanation from a probabilistic perspective. In the framework of likeliness, the explanation that maximizes the posterior probability $P(X|E_i)$ of the event $X$ explained by a certain explanation $E_i$ is a good explanation. Loveliness is an index defined from an axiological perspective. In the framework of loveliness,
\Erase{the goodness of explanation is mainly determined based on two indicators, simplicity and latent scope.}
\Add{studies have focused on simplicity and latent scope as indicators that determine the goodness of an explanation.}
Simplicity is an index showing the number of assumed causes; the smaller the number, the more preferable the explanation is to humans \cite{lombrozo}. 
Latent Scope is an event that is predicted to be caused by a cause but has not been observed as a result. 
The smaller the number of latent scope, the higher the estimation of posterior probability $P(X|E_i)$ \cite{harrypotter,Johnson}.

According to the indices of likeliness and loveliness, an autonomous agent's description of an approach to a goal must be composed of the minimum information necessary for comprehension. 
In this paper, we aimed to improve the likeliness of the explanation and reduce the latent scope by extracting the elements that are crucial to reaching the target state. 
We also aimed to improve the simplicity of the explanation by estimating the elements that are not understood by the user as discussed in the previous section. 

\subsection{Research on explainability}
\Add{
Research topics similar to those considered in this study include XAI, explainable AI planning (XAIP), and explainable reinforcement learning (XRL). 
}
The research areas are outlined in this section\Add{; refer to \cite{surveyXAR} for details.}

In the research area of XAI, several methods aimed at improving the explainability of classification models have been proposed. Typical examples are local interpretable model-agnostic explanations (LIME) \cite{lime}, Shapley Additive exPlanations (SHAP)\cite{shap}, and Anchors \cite{Anchors}, all of which improve the interpretability of the model by presenting the features that contributed to the prediction results. As mentioned above, these do not concern the explanation of the flow from the input features to the desired decision-making results. XAIP is a research area focused on transparency regarding system decision-making and planning. 
This emphasizes a balance between making predictable plans for users and explaining plans made by robots \cite{Chakraborti,XAIP2018WS,XAIP2019WS}. 
Majority of the research in this area considers the explainability of symbolic planning techniques; therefore, they are difficult to use directly with agents that learn autonomously. XRL is a research area that introduces the approach of XAI into the framework of reinforcement learning, and aims to interpret the learned policy in a form that can be understood by humans. Methods for learning the value of state-actions by reward source \cite{juozapaitis2019explainable}, methods replacing policies in the form of human-readable programs \cite{pirlICML18}, and methods extracting states in which the characteristics of policies are well represented \cite{HIGHLIGHTS} have been proposed. However, these methods are only aimed at increasing the transparency of the policy; they do not consider explaining the transition from the current state to the target state.

To solve the same problem as XAR, causal explanations based on causal inference \cite{Madumal2019ExplainableRL} and generation of explanations using counterfactual assumption \cite{Madumal2020DistalEF} have been proposed. However, in this framework, it is necessary for humans to create the causal structure of the environment, which makes it difficult to directly apply them to agents that learn autonomously. Because the proposed method can extract important elements for understanding the transition from the current state to the target state without explicitly providing the causal structure of the world, 
it becomes the first step towards bestowing the ability to explain on an agent that learns and makes decisions autonomously.

\Add{
\subsection{Contributions and limitations of this paper}
In this study, we propose a method for users to understand the path to the target state in the MDP by providing the minimum necessary information according to the above definition. 
The contributions of this study are as follows: $\rm(\hspace{.10em}i\hspace{.10em})$~proposal of a general-purpose method for explanation generation that is applicable to all autonomous agents based on the MDP framework, $\rm(\hspace{.10em}ii\hspace{.10em})$~proposal of a method that can present important elements for understanding the transition from the current state to the target state, without a human providing information in advance on the structure of the environment, and $\rm(\hspace{.10em}iii\hspace{.10em})$~proposal of a new explanation method that explicitly considers the user's understanding of events. }
\interfootnotelinepenalty=10000
\Add{
The limitations of this study are as follows. 
$\rm(\hspace{.10em}i\hspace{.10em})$~The environment must be represented by an MDP, 
$\rm(\hspace{.10em}ii\hspace{.10em})$~no other knowledge than the MDP is available, 
$\rm(\hspace{.10em}iii\hspace{.10em})$~the user and the robot share all the states of the MDP\footnote{The user and the robot share the states, but they do not share state transitions. This lack is the reason why an explanation is required, and is the main focus of this study.},
and 
$\rm(\hspace{.10em}iv\hspace{.10em})$~expressing the generated explanation in natural language is beyond the scope of this study. 
All of these are important problems in XAR and are discussed in \cite{surveyXAR}. 
}

\Erase{
\section*{Problem Formulation}
In this paper, we adopted the MDP framework for the agent's decision-making process to mathematically examine the explanations accepted by humans. In the MDP framework, the agent's decision-making is modeled using the following elements \cite{Madumal2020DistalEF}.
\begin{itemize}
    \item $\mathcal{A}$ ; Set of actions that the agent can select
    \item $\mathcal{S}$ ; Set of states
    \item $\mathcal{R}:\mathcal{S}\times \mathcal{A}\to \mathbb{R}$ ; Reward function
    \item $\mathcal{T}=\{P(s'\mid a,s)\}$ ; The probability of transition to the state $s'\in \mathcal{S}$ when taking action $a\in \mathcal{A}$ in a state  $s\in \mathcal{S}$
\end{itemize}
Here, the subset of the state-action pairs $(s,a)$ on an MDP is defined as explanation $E$. The purpose is to present the process of reaching the target state $s_{target}$ in a form that is acceptable to humans. To do so, as mentioned above, it is better to present the minimum information necessary to understand the path to the target state on the graph. That is, the set $E$ of state-action pairs that satisfy the following equation is obtained:
\begin{equation}
 \begin{split}
\label{definition}
&\mbox{$\min |E| ,\ subject\ to\  P(Reach(s_{target})\mid E)>\alpha$}
\end{split}
\end{equation}
Here, $Reach(s_{target})$ is the event of reaching the target state $s_{target}$, and $\alpha\in[0,1)$ is the level of comprehension required by the recipient of the explanation. Furthermore, $|E|$ is the number of elements in set $E$. 
To solve this optimization problem, it is necessary to design a clear derivation method of the questioner's level of understanding and the probability of reaching the target state.
However, a clear definition is exceedingly difficult. Therefore, in this study, we considered a heuristic solution in which $\rm(\hspace{.10em}i\hspace{.10em})$ the important elements for reaching the target state are identified, and $\rm(\hspace{.10em}ii\hspace{.10em})$ the elements that the user does not understand are presented. Each can be examined on an MDP as follows.
\begin{description}
    \item [$\rm(\hspace{.10em}i\hspace{.10em})$ Important elements for reaching the target state]\mbox{}\\
    According to the concepts of likeliness and latent scope mentioned above, an important element for reaching the target state is ``an element that is useful for imagining the process of reaching the target state and is absolutely necessary when reaching the target state.'' Therefore, ``a state-action pair in which it is difficult to find an alternative route to the target state on the MDP'' can be defined as an important element.
    \item [$\rm(\hspace{.10em}ii\hspace{.10em})$ Elements that the user does not understand]\mbox{}\\
    If the user understands all the transitions to the target state on the MDP, s/he can understand the process through which each action leads toward the target. Based on this, the elements that the user does not understand are ``elements for which the user cannot correctly estimate the transition on the MDP.''
\end{description}
In this study, we proposed a method for users to understand the path to the target state on the MDP by providing the minimum necessary information according to the above definition. The contributions of this study are as follows. $\rm(\hspace{.10em}i\hspace{.10em})$ Proposal of a general-purpose explanation generation method that is applicable to all autonomous agents based on the MDP framework, $\rm(\hspace{.10em}ii\hspace{.10em})$ proposal of a method that can present important elements for understanding the transition from the current state to the target state, without a human providing information on the structure of the environment in advance, and $\rm(\hspace{.10em}iii\hspace{.10em})$ proposal of a new explanation method that explicitly considers information on the user's understanding of events.
}

\section{Proposed Method}
\label{chap:提案手法}
In this study, an explanation is generated in three steps, to present the minimum information necessary for humans to understand the autonomous agent's progress to the goal.
 \begin{description}
  \item [Learning the world model]\mbox{}\\A predictive model of the world (world model) that can estimate state transitions without action in the real environment is learned. This model is used to extract important state-action pairs to reach the target state and estimate the user's understanding of events.
  \item [Identification of subgoals based on calculation of importance]\mbox{}\\The importance is calculated using the world model, and subgoals (state-action pairs for which it is difficult to find an alternative route to the target state on the MDP) are extracted.
  \item [Identification of key points and presentation of explanation]\mbox{}\\The elements on the MDP that are not understood by the user are estimated using the world model and used for explanation.
 \end{description}
\Erase{
The concept of explanation assumed in this study is shown in Figure \ref{fig:intro}. 
For example, when it is essential to ``pick up the key'' and ``open the door'' to reach the target state, the explanation of ``it is necessary to first pick up the key'' is given to a questioner who thinks that the door should be opened directly. 
}
It should be noted that it was assumed that the questioner and agent represent the world using the same set of states in this study. 
However, it is quite possible that the questioner and agent use different states to represent the world. It should be noted that to apply the proposed method in such a case, it is necessary to convert the state space into a representation using a common set of states.

 \begin{figure}[t]
  \begin{center}
    \includegraphics[width=0.8\linewidth]{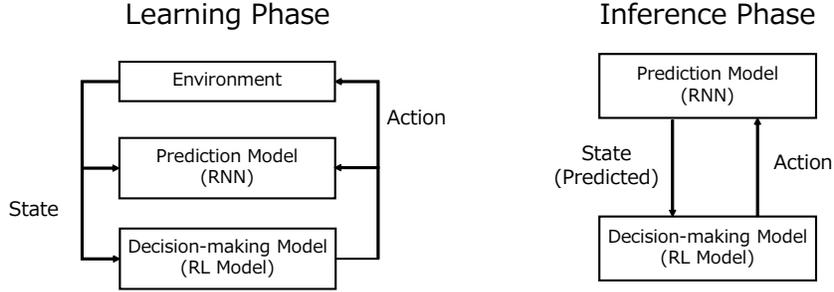}
    \caption[Relationship between the prediction model and action decision model]{Relationship between the prediction model and action decision model. During learning, observation information from the environment and action information from the action decision model are input to the prediction model for learning. Furthermore, during inference, the observation information series can be retrieved using a prediction model without using observation information from the environment.
    }
    \label{fig:worldmodel} 
  \end{center}
\end{figure}
 
\subsection{Learning the world model}
To extract the important elements and respond to various questions, the agent must learn the correspondence between its actions and environmental changes. That is, we need a prediction model that enables the counterfactual inference that ``if action $a$ is taken, the environment will become $s$.'' Ha {\it et al.} proposed a ``world model'' \cite{NIPS2018_7512} for learning such a predictive model of the world. In the world model, the correspondence between the actions of the autonomous agent and changes in the outside world can be learned using a recurrent neural network (RNN), and the model can be used for acquiring an action policy for the agent and determining actions.

In this study, we built on the idea of the world model by implementing it on the RNN. Let the input of the model be the observation $s_t$ and action $a_t$, of the agent and the output be the observation $s_{t+1}$ for the next point in time. For the agent to acquire the goal-oriented policy, the learning data is acquired at the same time as the reinforcement learning. Because a complete observation is assumed here, the state and observation are expressed equally as $s_t$. The world model makes it possible to restore the state-action sequence without using observation information from the environment during the counterfactual inference (Figure \ref{fig:worldmodel}). In this study, the policy and world model are acquired simultaneously; however, this does not necessarily have to be by simultaneous learning. 
What is needed here is to model the relationship between the action and state transition. 
Then the actions based on the policy can be simulated using the model.

 \subsection{Identification of subgoals based on calculation of importance}
\begin{figure}[t]
  \begin{center} 
    \includegraphics[width=0.8\linewidth]{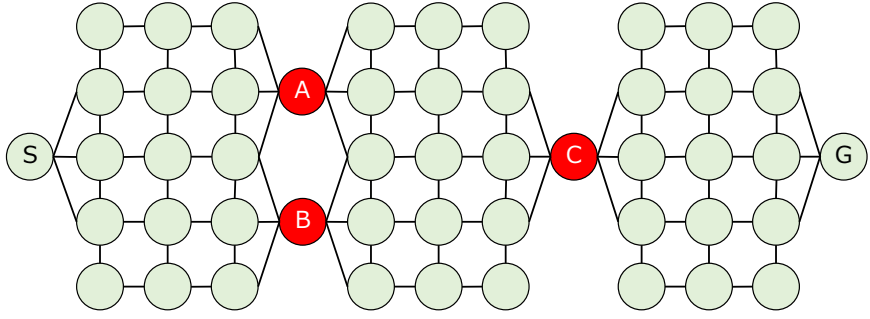}
    \caption[Conceptual diagram of importance and subgoals]{Conceptual diagram of importance and subgoals. When considering a route to goal G from start point S, the process will always pass through Node C, and A or B. Therefore, the most important node is C, followed by A and B.
    }
    \label{fig:subgoal} 
  \end{center}
\end{figure}

To provide the minimum information necessary to understand the approach to achieving the goal, it is necessary to identify the important elements in the series. In this study, the difficulty of finding an alternative series through which it is possible to move to the target state in the MDP environment is defined as the importance, and an important element is made a subgoal (Figure \ref{fig:subgoal}).
  
 \begin{figure}[t]
  \centering 
    \includegraphics[width=0.8\linewidth]{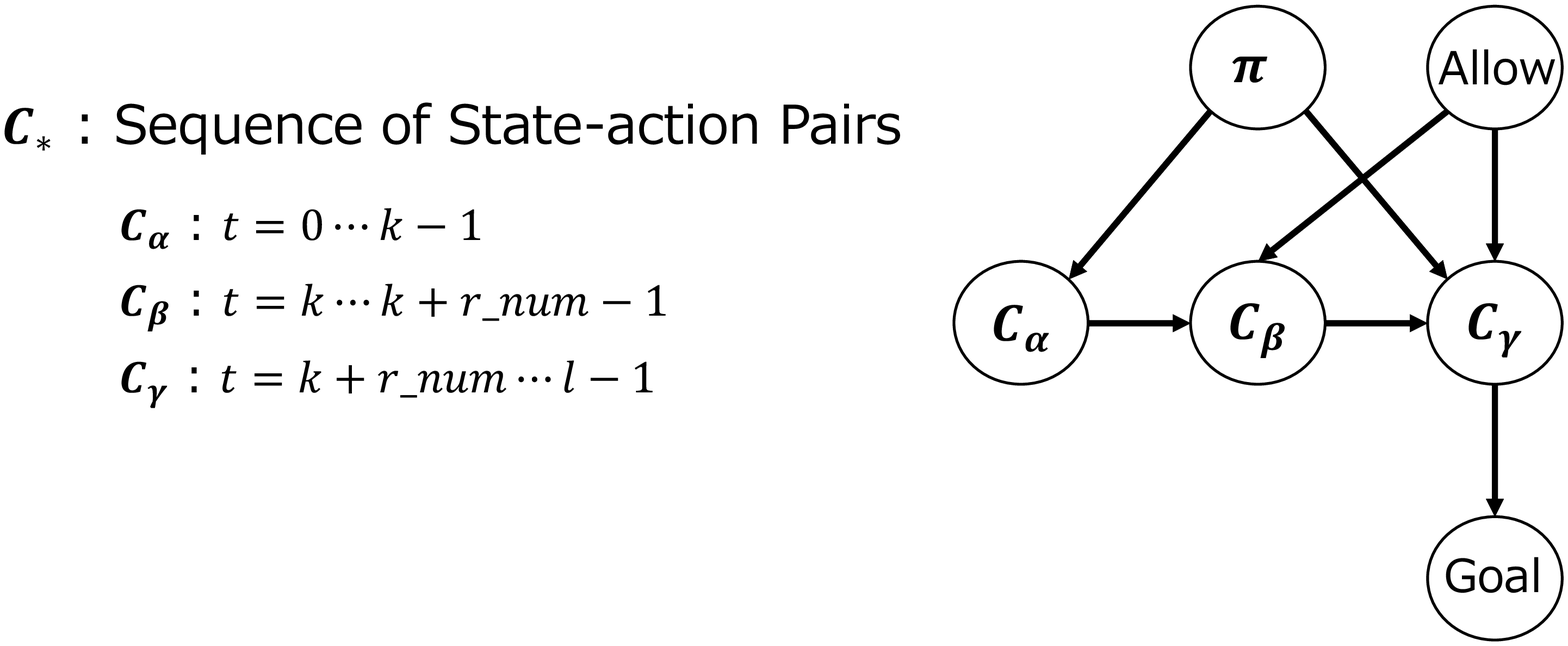}
    \caption[Cause-effect diagram]{
    Cause-effect diagram of variables considered in the formulation using ACE. Exogenous variables are omitted. Because the variable $allow$ that intervenes in this case does not have parents, the intervention diagram is also the same.
    }
    \label{fig:inga} 
\end{figure}

\subsubsection{Calculation of importance}
Let {\boldmath $S_{opt}$} be the set of all the states on the optimal path from the initial state $s_{0}$ to the target state $s_g$. The difficulty in a certain state $s_{f}\in$ {\boldmath $S_{opt}$} of finding an alternative series through which it is possible to move to the target state is defined as ``the probability of returning to $s_{f}$ again when acting according to action policy $\pi$ after acting randomly from $s_{f}$''. This probability can be estimated through iterative calculation using the world model. Here, the probability of returning to this $s_{f}$ is defined as the importance $I(s_f, \pi)$ in policy $\pi$ of state $s_{f}$.
Algorithm \ref{alg1} shows the method of calculating importance.

Importance, as defined here, is equivalent to investigating how directly the cause of the optimal action in the state relates to the result of the goal. Therefore, this probability value can be formulated using the following equation as the average causal effect (ACE) \cite{Pearl}, defined by a framework of statistical causal reasoning.
%
\begin{equation}
 \begin{split}
 \mbox{ACE}
 &=P(Goal=1\ |\ \mbox{$\pi$},\mbox{$s_f$},do(allow=1))-P(Goal=1\ |\ \mbox{$\pi$},\mbox{$s_f$},do(allow=0))\\
 &=1-P(Goal=1\ |\ \mbox{$\pi$},\mbox{$s_f$},do(allow=0)),
 \end{split}
 \end{equation}
where $do(\cdot)$ is a Pearl do operator \cite{Pearl} that indicates intervention. Figure \ref{fig:inga} is a cause-effect diagram of the variables used in the formulation. Here, let $s_{f}$ be the state for which the importance is to be obtained and $s_{g}$ be the target state; then, $k$, $r\_num$, and $l$ are the time taken to reach $s_{f}$, the number of random actions, and the time to reach $s_{g}$, respectively. Let the time at the position of the initial state $s_{0}$ be $t=0$. If the state-action sequence {\boldmath $C_x$} is defined as {\boldmath $C_x=(s_x,a_x)$}, the action sequence {\boldmath $a_\alpha$}, {\boldmath $a_\gamma$} is determined by the policy, and {\boldmath $a_\beta$} is randomly determined based on the selectable actions. Let an action that can be selected refer to ``action $a_t$, where $s_t\neq s_{t+1}$'', and taking an action whereby the episode is completed be prohibited. When $allow=0$, it is also prohibited to take the optimum action $a_{p}$ at $s_{f}$.
The variable $Goal$ becomes 1, when the target state can be reached and 0, otherwise. The variable $allow$ becomes 1 when taking the optimal action $a_{p}$ at $s_{f}$ and 0, otherwise.
From the formulation above, the importance can be interpreted as the effect of ``being able to select $a_{p}$ at $s_{f}$'' for the event of ``reaching the target state'' in the state-action sequence {\boldmath $C^*$}= ({\boldmath $C_\alpha$},\ {\boldmath $C_\beta$},\ {\boldmath $C_\gamma$}).
The importance $I(s_f, \pi)$ calculated using Algorithm \ref{alg1} can be interpreted as an approximation of the ACE.

\begin{algorithm}[t]                      
    \caption{Calculate importance $I(s_f, \pi)$}         
    \label{alg1}                          
    \begin{algorithmic}
        \Require~~\\
        $r\_num\gets$ number of random actions per trial\\
        $s\_num\gets$ number of trials\\
        $s_f\gets$ featured state\\
        $s_g\gets$ goal state
        \Ensure{$I(s_f, \pi)$}
        \State $keep\_imp\gets 0$
        \For{$i\ {\bf in}\ s\_num$}
        \State $s_{cf}\gets$ state after $r\_num$ random actions
        \While{$s_{cf} \neq s_{g}$}
        \State $a_{p}\gets$ optimal action at $s_{cf}$
        \State $s_{cf}\gets$ predicted state after $a_{p}$
        \If{$s_{cf}=s_{f}$}
            \State $keep\_imp\gets keep\_imp+1$
            \State {\bf break}
        \EndIf
        \EndWhile
        \EndFor
        \State $I(s_f, \pi)\ =\ keep\_imp/s\_num$\\
        \Return $I(s_f, \pi)$
    \end{algorithmic}
\end{algorithm}

\subsubsection{Extraction of subgoals using importance}
Using the importance of each state obtained, the subgoal $S$ in the state-action sequence ({\boldmath $s$},\ {\boldmath $a$}) for policy $\pi$ is calculated using Equation (\ref{subgoal}).
\begin{equation}
 \label{subgoal}
 S[ (\mbox{\boldmath $s$},\mbox{\boldmath $a$}), \pi]
 =\{{(s,a)}_t\ |\ {(s,a)}_t\in{(\mbox{\boldmath $s$},\mbox{\boldmath $a$})},~I(s_t,\pi)>\varepsilon,~I(s_{t+1},\pi)<\varepsilon \}
\end{equation}
$\varepsilon$ is the threshold of the importance; although it is set empirically, in the experiment described below, it is the average value of the total. From this equation, in the state-action sequence ({\boldmath $s$},\ {\boldmath $a$}), the state-action pair ${(s,a)}_t$ in which the importance of $s_t$ is higher than $\varepsilon$ and the importance of $s_{t+1}$ is lower than $\varepsilon$ is extracted as a subgoal. 
To avoid extracting the vicinity of the subgoal in addition to the original important element, the subgoal is set not to extract adjacent state-action pairs on the series.

\subsection{Identification of key points and presentation of explanation}
To satisfy the requirement of simplicity, it is desirable not only to identify subgoals but also to explain them according to the understanding of the user. In other words, it is required to estimate the information (key points) that the information recipient lacks from the element (subgoal $S$) that is important for reaching the target state, and to generate a more concise explanation. The procedure for estimating the key points and generating explanations is given below, and the concept is shown in Figure \ref{fig:keypoint_image}.

\begin{enumerate}
  \item For the action sequence ${\mbox{\boldmath $a$}}^{bad}$
  \Add{given by the questioner}
  that does not lead to the target state,
  \Erase{given by the questioner}the state transition is estimated using the world model, and the state-action sequence ${(\mbox{\boldmath $s$},\mbox{\boldmath $a$})}^{bad}$ is generated.
  \item The key point is the state-action pair ${(s,a)}_{t}$ that is not included in ${(\mbox{\boldmath $s$},\mbox{\boldmath $a$})}^{bad}$, and included in the subgoal of the state-action sequence ${(\mbox{\boldmath $s$},\mbox{\boldmath $a$})}^{good}$ that can lead to the target state.
  Furthermore, to present the minimum information, the element with the smallest value of $t$ in ${(s,a)}_{t}$ is selected and made the explanation.
\end{enumerate}
  \begin{figure}[t]
  \centering
    \includegraphics[width=1.0\linewidth]{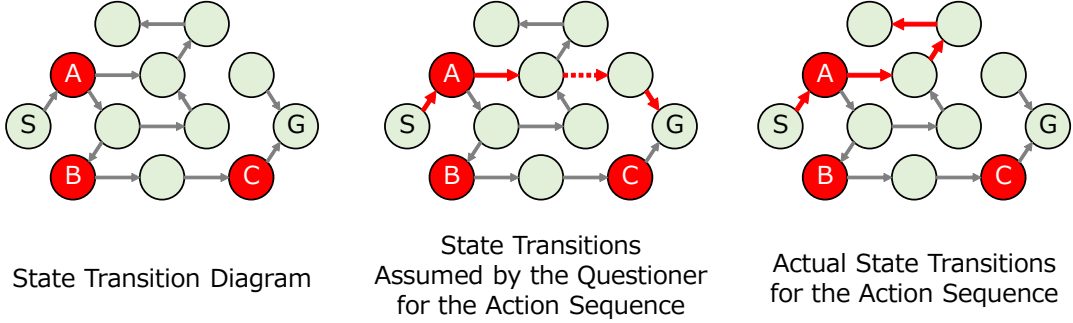}
    \caption[Conceptual diagram of key point estimation and explanation generation]{
    Conceptual diagram of key point estimation and explanation generation. Suppose that the questioner assumes an erroneous state transition and gives an action sequence ${\mbox{\boldmath $a$}}^{bad}$ that does not lead to the target state as a question.
    The agent estimates the actual state transition when following the action sequence, and the key point is the state that cannot be passed among the subgoals. 
    Furthermore, the first state to be reached among the key points is presented as an explanation. Assuming that Nodes A, B, and C in the figure are subgoals, B and C are the key points, and B is the explanation.}
    
    \label{fig:keypoint_image} 
\end{figure}
The formula for deriving the key points is as follows.
\begin{equation}
\label{keypoint}
\mbox{keypoints}\left({(\mbox{\boldmath $s$},\mbox{\boldmath $a$})}^{bad}, S[{(\mbox{\boldmath $s$},\mbox{\boldmath $a$})}^{good}, \pi]\right)
=\{{(s,a)}_t\ |\ {(s,a)}_t\in S[{(\mbox{\boldmath $s$},\mbox{\boldmath $a$})}^{good}, \pi]\backslash{(\mbox{\boldmath $s$},\mbox{\boldmath $a$})}^{bad}\}
\end{equation}
The explanation can also be described using Equation (\ref{explanation}).
 \begin{equation}
\label{explanation}
\mbox{explanation(keypoints)}
=\{{(s,a)}_{t_{min}}\in \mbox{keypoints}\ |\ \forall{(s,a)}_{t}\in \mbox{keypoints},\ t\geq t_{min}\}
\end{equation}

\section{Simulation Experiment}
\label{chap:実験}
\begin{figure}[t]
\centering
    \includegraphics[width=0.7\linewidth]{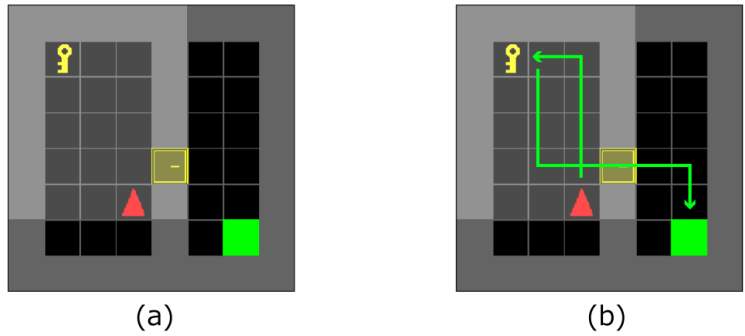}
    \caption{ (a) Experimental environment (b) Optimal path}
    \label{fig:env} 
\end{figure}

We applied the proposed method in a simulation environment to determine whether the important elements for grasping the transition from the current state to the target state can be extracted as explanatory factors. As an abstraction of the agent's MDP in the experimental environment, we used a partially modified grid environment \cite{gym_minigrid} in which multiple objects were placed (Figure \ref{fig:env}). In this environment, the agent (red) obtains the key, opens the yellow door, and reaches the green goal to receive a reward (maximum reward: 1; decreases over time). There are five types of agent actions: go straight, face left, face right, take key on the forward grid, and open/close door. Furthermore, the agent observes a total of five dimensions: its own absolute position ($x$ coordinate, $y$ coordinate) and orientation, possession/non-possession of key, and door open/closed.

 \begin{figure}[t]
  \centering
    \includegraphics[width=0.8\linewidth]{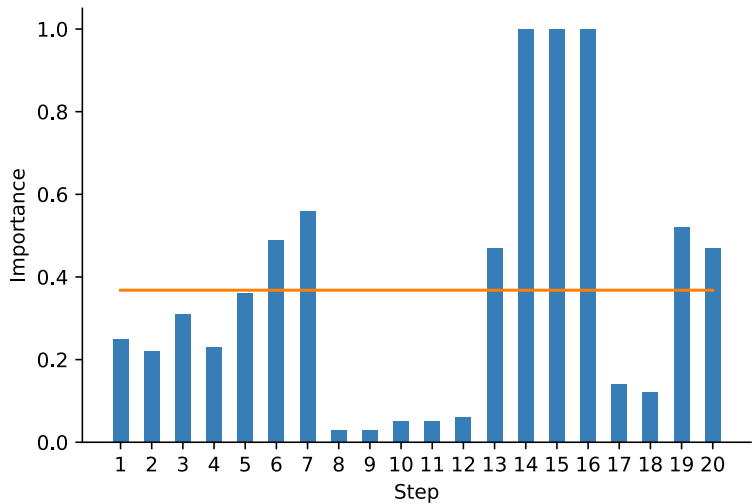}
    \caption{Importance of each state on the optimal path}
    \label{fig:imp1}
\end{figure}
 \begin{figure}[t]
  \centering 
    \includegraphics[width=0.9\linewidth]{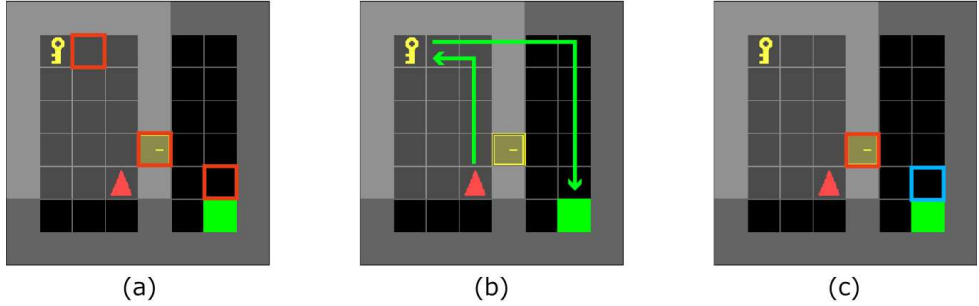}
    \caption[Derivation result for the optimal route]{Derivation result for the optimal route. (a) Subgoals (b) Route assumed by the questioner (c) Key points (red box, blue box) and explanation (red box) for route assumed by the questioner
    }
    \label{fig:keypoint1}
\end{figure}

 \subsection{Results of the proposed method}
In the environment described above, the proposed algorithm for extracting the important elements is applied to an agent that learned the action policy by deep Q-network (DQN) \cite{DQN} and the world model through simple RNN. The observation information and actions converted into one-hot vectors, rather than the image information, are input to the networks. In addition, the output of the DQN is the action value of each possible action, and the output of the RNN is the probability value of the observation information for the next time step. Learning was performed until the loss values of both networks converged.

 \subsubsection{Importance derivation results}
 The importance derived in each state of the optimal path in the environment depicted in Figure \ref{fig:env} is shown in Figure \ref{fig:imp1}. The orange line represents the average value of importance. When considering the results, it should be noted that the agent can only move in the direction it is facing. For example, when an agent facing up moves one square to the left, two steps of “turn left → go straight`` are required as actions.
Consequently, it was confirmed that the importance was high near ``get key (Step 7)'', ``pass through door (Steps 14–-16)'', and ``reach goal (Step 20)'', which are highly necessary states to pass through.

\subsubsection{Subgoal, key point, and explanation identification results}
\label{3.1.2}
The subgoals estimated using the importance and the key points and explanation for the question indicating misunderstanding of the location of the door are shown in Figure \ref{fig:keypoint1}\footnote{\Add{When following the route in Figure \ref{fig:keypoint1}(b), the action sequence ${\mbox{\boldmath $a$}}^{bad}$ is given as a query. In this experiment, it is assumed that the representation of the state space is shared between the agent and the questioner. That is, their perceptions differ only regarding state transitions related to the position of the door.}}.

The threshold value $\varepsilon$ of the subgoal is the average value of the importance of all the steps on the optimum path; the importance in the target state was set to zero because it could not be calculated by definition. Consequently, ``get key'', ``pass through door'', and ``reach goal'' were extracted as the subgoals, and ``pass through door'' was extracted as the explanation. 
It means that the elements with few alternative routes were extracted as the subgoals, and it is also possible to present a state-action pair that the questioner has not reached.

\subsection{Evaluation of subgoal usefulness}
\begin{figure}[t]
    \centering 
      \includegraphics[width=0.8\linewidth]{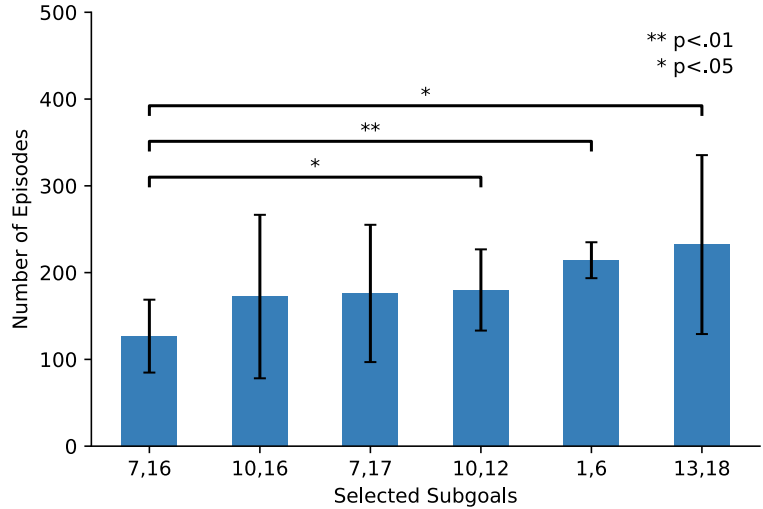}
      \caption{Comparison of the number of episodes required to acquire a policy according to the presented subgoals} 
      \label{fig:select_subgoal} 
  \end{figure}

We verified that the selected subgoals were useful for understanding the approach to reaching the goal from the perspective of the time required to acquire the policy of the agent. 
In this experiment, an agent that had not learned the policy was used, which was different from the agent that generated the explanation. 
In the environment shown in Figure \ref{fig:env}, the average value of the required number of episodes when the optimal policy was acquired ten times each was calculated for the case where a sub-reward (maximum reward value: 0.5; decreased over time) was added to subgoals randomly selected from the optimal state-action sequence and subgoals derived using the proposed method (Steps 7 and 16, excluding the original target state). 
Furthermore, an unpaired two-sample t-test was performed at the significance level $\alpha=.05$ for the derived subgoal and each random subgoal. The results are shown in Figure \ref{fig:select_subgoal}. In this verification, the initial position of the agent was fixed, and learning was completed when the optimum policy was obtained from the initial position. It was confirmed that there was no significant difference when one of the two subgoals was common to the derived subgoal. However, in \ref{3.1.2}, a significant difference was found when both subgoals were different from the subgoal. From the foregoing, it could be inferred that the derived subgoal was optimal, with respect to each random subgoal.

\subsection{Validity evaluation of the number of subgoals}
\begin{figure}[t]
    \centering
      \includegraphics[width=0.8\linewidth]{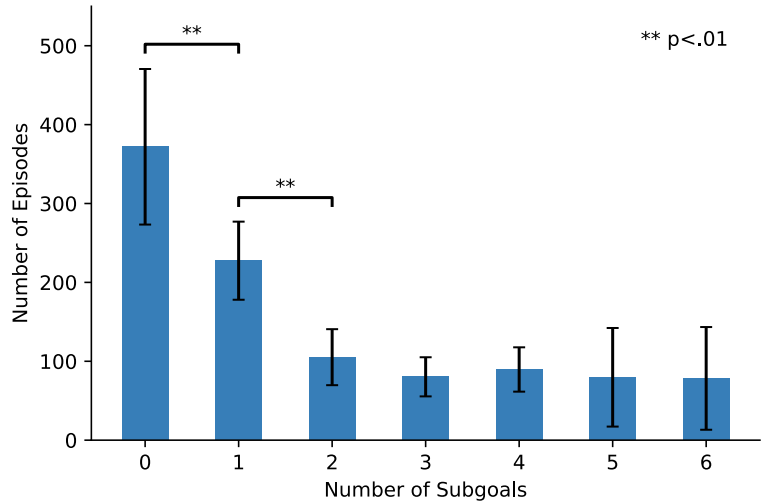}
      \caption{Comparison of the number of episodes required to acquire a policy according to the presented subgoals} 
      \label{fig:subgoal_num} 
  \end{figure}

We verified that the derived number of subgoals was the minimum number of elements necessary to sufficiently understand the approach to the goal. In this experiment, the agent used had not learned the policy and was different from the one that generated the explanation. An additional subgoal $s_{add}$ was added to the derived subgoals (excluding the original target state), and the relationship between the total number of subgoals and the number of episodes required to acquire the optimal policy was investigated. Sub-rewards (maximum reward value: 0.5; decreased over time) were allotted to the subgoals and additional subgoals. The same verification was performed when the subgoals were fewer than the derived subgoals. The following two requirements were established for $s_{add}$: 
\begin{itemize}
    \item When considering the shortest path, the reward was higher when reaching the next subgoal $s_{next}$ of $s_{add}$ than when reaching $s_{add}$ again after passing it once.
    \item $s_{add}$ divided the maximum distance between existing subgoals into two.
\end{itemize}
The verification results for the environment shown in Figure \ref{fig:env} are shown in Figure \ref{fig:subgoal_num}. There was a significant difference only between the distributions of 0 and 1, and 1 and 2 subgoals. Furthermore, even if the number of subgoals increased, the number of episodes required to learn the optimal policy did not change significantly. Therefore, it was deduced that the subgoal of ``the minimum number of elements necessary for sufficiently understanding the event'' based on the likeliness and loveliness can be extracted using the proposed method.

\section{Subject Experiments}
In the same environment as the simulation experiment, we generated an explanation on the action of the agent and verified whether the explanation enabled the user to understand the transition from the current state to the target state. The subjects of the experiment were a total of 60 subjects, 24 males and 36 females, aged 16 to 54 years old. Prior to the experiment, the following three points of information on the grid environment were given to the subjects.
\begin{itemize}
    \item The agent heads for the goal.
    \item It has to pick up the key, then go through the door to reach the goal.
    \item The key and door positions change each time.
\end{itemize}

\begin{figure}[t]
    \centering
      \includegraphics[width=1.0\linewidth]{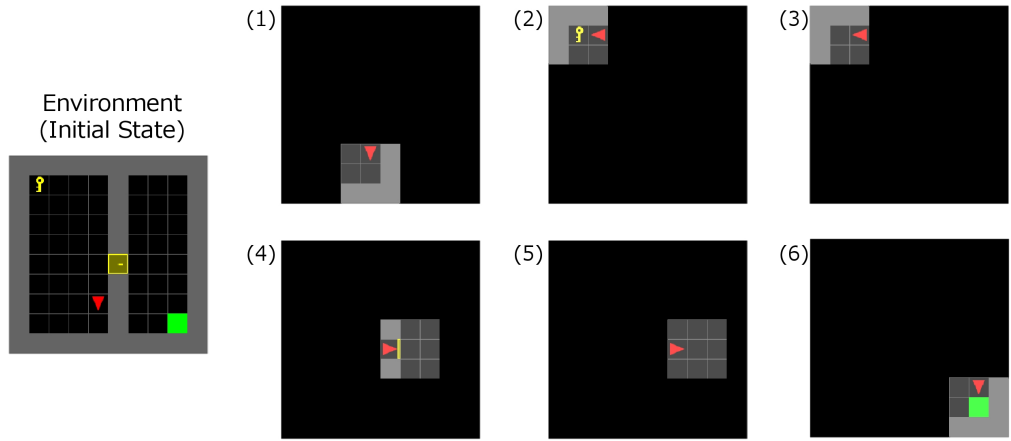}
      \caption{Example of images presented during the subject experiment} 
      \label{fig:sub_experiment_image}
\end{figure}

In this experiment, the state-action pair was expressed by showing the transition from the state presented as an explanation to the state at the next time step. When the grid environment was presented to the subject, an image in which only the $3\times3$ squares in front of the agent could be observed was always presented. This was to re-create the situation in which humans do not have a bird's eye view of all the elements on the MDP when attempting to understand the agent's decision-making process. The environment used in the experiment and an example of the images presented during the explanation using the subgoals are shown in Figure \ref{fig:sub_experiment_image}. For example, Figure \ref{fig:sub_experiment_image} shows (1) the initial state; (2) and (3) getting the key; (4) and (5) passing through the door; and (6) reaching the goal.

\begin{figure}[t]
    \centering 
      \includegraphics[width=0.8\linewidth]{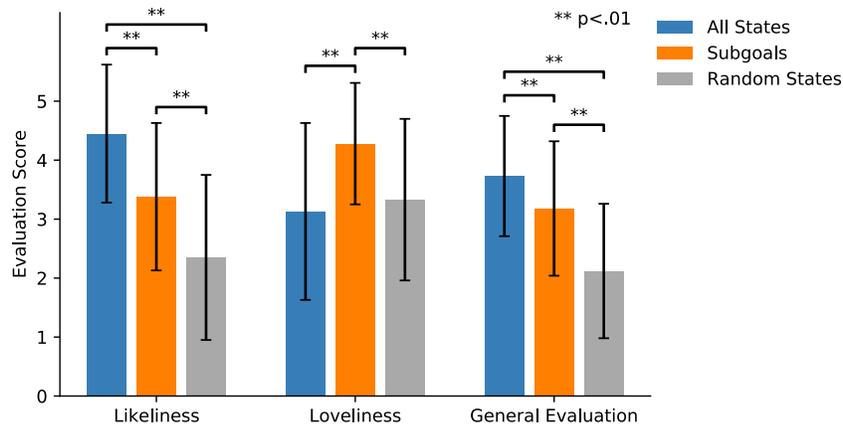}
      \caption{Results of Subject Experiment 1}
      \label{fig:experiment1} 
  \end{figure}

\subsection{Experiment 1}
We verified whether the presentation of subgoals sufficed as an explanation on the transition from the current state to the target state. First, the subject was given the current position of the agent. At that point, the subject did not know the positions of the key and the door. Subsequently, the agent gave the subject one of the following information: (a) all the state-action pairs that it will pass, (b) a state-action pair extracted as a subgoal, or (c) a state-action pair randomly selected from the optimal path. Finally, the subject evaluated the given explanation based on the following three items.
\begin{enumerate}[$\rm(\hspace{.10em}i\hspace{.10em})$]
\item Could you predict the route that the agent took? (likeliness evaluation)
\item Was it a succinct explanation? (loveliness evaluation)
\item Comprehensive evaluation of explanation (general evaluation)
\end{enumerate}

Three environments with different key and door positions were used in the experiment; the environments appeared in the same order for all subjects. The description method used for each environment was determined randomly. However, the experiment was designed such that one description based on each description method was given to each subject.

The results are shown in Figure \ref{fig:experiment1}. For all the evaluation items, the presentation of the subgoals derived using the proposed method was highly evaluated, compared to the presentation of random state-action pairs. In addition, in the loveliness evaluation, the presentation of the subgoals was highly evaluated, compared to the presentation of the entire series. It is also interesting that the correlation coefficient between the likeliness and general evaluation was high, at r = 0.74, whereas the correlation coefficient between the loveliness and general evaluation was very low, at r = 0.08. In this experiment, one of the evaluation items was ``whether the route could be predicted;'' therefore, the subjects possibly considered being able to predict the route as an absolute condition of the explanation, and the quality of the explanation did not reach a level that considered an evaluation of loveliness.

Here, we hypothesized that the effect of loveliness on the general evaluation of the explanation differed for the low (Likeliness = 1, 2, 3) and high likeliness class (Likeliness = 4, 5). We performed a two-factor analysis of variance (ANOVA), based on which a main effect was observed for likeliness ($F(3.98)= 67.65,p<.001$), but not for loveliness ($F(3.98)=2.20,p=.143$).
Further, it was confirmed that the interaction was significant ($F(3.98)=6.56,p=.012$). Based on the observed interaction, the group was divided into a likeliness = 1, 2, 3 group and a likeliness = 4, 5 group, and a simple main effect analysis was performed. The results are shown in Table \ref{table:tanjyun}. Regarding the p-value and the significant difference, the results of an unpaired one-sided t-test at a significance level $\alpha=.05$ for two distributions in the same likeliness class are shown. The other items in the table show the average value of the general evaluation under each condition.

Following the analysis, the simple main effect of loveliness was not observed in the class with low likeliness (Likeliness = 1, 2, 3), whereas it was observed in the class with high likeliness (Likeliness = 4, 5). The above demonstrates that the results of the experiments validated the hypothesis that the effect of loveliness on the general evaluation of the explanation differed between the low and high likeliness classes is valid in the results of this experiment. 
From the results of the two-factor ANOVA and the fact that loveliness of the subgoals derived using the proposed method were high, it is suggested that the proposed method could generate user-friendly explanations when the user does not require a detailed understanding of the transition from the current state to the target state.

Experiment 1 revealed that presenting the subgoals extracted by the proposed method was a better explanation than presenting random elements on the optimal path. When the user wants to understand the entire process from the current state to the target state, the whole series should be presented. However, there are several situations in which the user does not require detailed information, wherein the subgoals derived by the proposed method suffice.
\Add{Therefore, an important question for future research is how to determine the type of explanation the user expects (specifically, whether the user wants to know the details or just the outline of the series). It is expected that more user-friendly explanations will be made possible with proper use of the explanations generated by the proposed method and the explanations in the whole series presentation.}

\begin{table}[t]
\begin{center}
\caption{Results of simple main effect analysis}
\scalebox{1.0}{
\begin{tabular}{c|c c | c c}
   \diagbox{Likeliness}{Loveliness}  & 1,2,3  &  4,5 & $p$-value	 & Significant difference \\\hline 
1,2,3   & 2.17  & 1.98 & $p=.251$ & None \\
4,5   & 3.23  & 4.07  & $p<.001$ & Present \\
\end{tabular}
}
\label{table:tanjyun}
\end{center}
\end{table}

\subsection{Experiment 2}
In Experiment 2, we verified whether the presentation of key points was sufficient to explain the reason for the agent's current action; we also verified whether the presentation of the first key point was sufficient. First, the subject was presented with a video of the actions that had been taken by the agent and a text of the actions to be taken. The agent then presented the subject with either (a) all the key points or (b) only the first key point. Finally, the subject evaluated the given explanation based on the following three items.
\begin{enumerate}[$\rm(\hspace{.10em}i\hspace{.10em})$]
\item Did you understand the reason for the next action taken by the agent? (likeliness evaluation)
\item Was it a succinct explanation? (loveliness evaluation)
\item Comprehensive evaluation of explanation (general evaluation)
\end{enumerate}
In the experiment, two environments with different key and door positions were used. \Add{The presented environment and explanation were determined in the same way as in Experiment 1.}
\Erase{each of which appeared in the same order for all the subjects, were used. Although the explanation method used for both environments was determined randomly, the experiment was designed such that one explanation based on each explanation method was given to each subject.}

\begin{figure}[t]
    \centering 
      \includegraphics[width=0.8\linewidth]{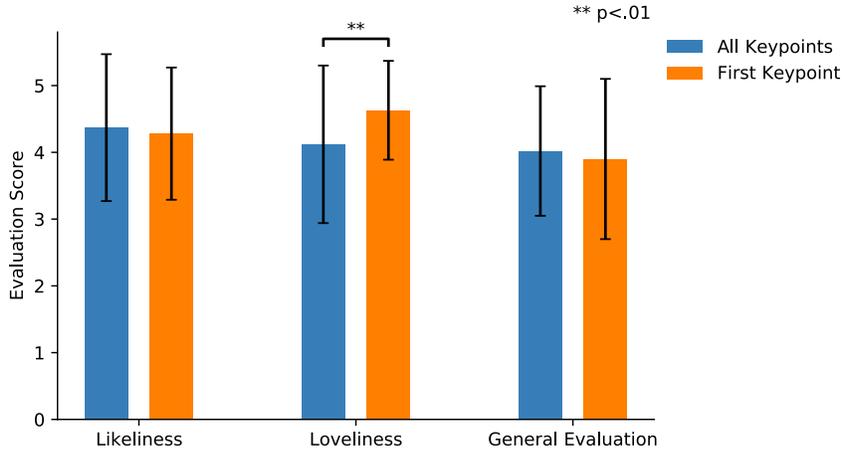}
      \caption{\small Results of Subject Experiment 2} 
      \label{fig:experiment2} 
  \end{figure}

The results are shown in Figure \ref{fig:experiment2}. 
\Add{The likeliness score obtained in Experiment 2 is higher than that obtained when the subgoal was presented in Experiment 1.
This difference arises because the questions used to evaluate the likeliness are different.
Considering only the reason for the next action taken by the agent, the explanation generated by the proposed method was able to obtain likeliness scores of over 4.
However, for route prediction, many subjects seemed to want to know the details of the route taken by the agent, and the resulting likeliness score was 3.38.}
 \Erase{the case where all the key points were presented and the one where the first key point was presented, evaluations equivalent to the highest evaluation in Experiment 1 were obtained for all the question items.}
 Further, \Add{only loveliness differed significantly in the evaluation of the two explanation methods.} 
 Comparing the variance of the general evaluation, $s^2=0.93$, when all the key points were presented, and $s^2=1.45$, when the first key point was presented. From this, it was deduced that the presentation of the first key point was an explanation method with a large degree of individual preference. In other words, the presentation of the first key point was a particularly good explanation for those who were content knowing merely knowing that the key was necessary (others desired to know what happened after the key was obtained).

\section{Discussion}
\label{chap:考察}
\subsection{Building the world model \Add{in the real world}}
In this study, the observation information was directly treated as a single state and a prediction model was generated. However, in the real world, observation information does not take discrete values; therefore, it cannot be directly represented by an MDP.
Our method is a method of presenting state-action pairs on MDP, so one important work is to build a world model that could be interpreted for real-world problems.
\Add{
For example, Zhang et al. \cite{zhang} proposed a method for building a discrete world model for agents that receive continuous features as observations.
Using this method for the latent representation obtained by the recurrent network makes it possible to apply our method to agents that observe partial and continuous features.}

\Add{
In a real environment, it is possible to use sensors equipped in the environment or in the robots. However, observations obtained in this manner do not always represent the correct environmental state.
Although beyond the scope of this study, proper estimation of environmental conditions from sensor observations is also an important issue.
}

\subsection{\Add{Use of a hierarchical world model}}
An advantage of this method is that it is applicable to any world model.
The granularity of the explanations presented can be modified by manipulating the information granularity of the world model. For example, although taking the key from the left or right are different actions at a lower level, they are considered the same action, taking the key, at a higher level. Modifying this granularity makes it possible to provide information that the user really wants. If the agent has world models of various granularity in advance, it may be possible to generate explanations that are better suited to the user's request.

Another advantage of this method is that it can generate explanations without knowledge of the details of the world model. Humans provide higher-level conceptual explanations, rather than ones based on specific observation information in cases where significant information is missing. Similarly, this method can continue to generate explanations using the same algorithm by shifting to inference on a coarser-grained world model when detailed information is missing.

\Add{
Our method of calculating importance assumes that the agent's world model can accurately represent the environment.
If the agent's world model is not well trained, it behaves differently from the real environment and accurate importance cannot be calculated even if the trained policy is used.
This problem can also be solved by generating explanations using a hierarchical and abstracted world model.
}

\subsection{Approach to deriving key elements}
The proposed importance derivation algorithm performs iterative calculations for each state. Therefore, if the state space is extensive and an alternative route search in the vicinity is inadequate, the calculation volume becomes extremely large.
\Add{This computational complexity problem is a major constraint on the use of our method in real-world problems.}

However, because this algorithm performs all operations on a prediction model without using the actual environment, parallel calculation is possible, and compression of the calculation time can be expected. Furthermore, the presentation of subgoals may not suffice as an explanation in an extensive state space. Therefore, even if a few subgoals are presented in an extensive search space, the distance between the subgoals may be so large that it impedes the user's comprehension of the state transition.
In other words, as an approach to compress the calculation volumes, deploying the hierarchical world model described above might be effective. By adjusting the granularity of the world model according to the required explanation, the calculation volume can be reduced and the granularity of the explanation can be adjusted.

In this study, the importance of each state was calculated using the probability of reaching the target state; however, the importance can also be calculated using the expected value of the reward. 
Because of importance calculation method, in an environment with few state loops (e.g. a highway), importance values of all the state-action pairs take small values in our method 
(e.g., although the decision of action at highway interchanges are important, they are not extracted as important because they exit the interchange during a random search). In such an environment, the important elements can be extracted based on the difference between the expected value of the reward when the attention action is permitted and when it is not. This is equivalent to seeking a causal effect for the expected reward value for the action being taken.

\subsection{User-specific explanation}
In this study, the state sequence was retrieved on the world model of the agent based on the action sequence assumed by the user, and the explanation was generated. However, in real problems, the precise action sequence assumed by the user is not known, and even if the action sequence can be estimated, the level of explanation required is user-dependent, as described in the section on subject experiments. Therefore, it is necessary to estimate the user's current level of understanding and the required level of explanation with greater accuracy. The hypothesis proposed in this paper is that estimating the state that can be attained by the user on the MDP is the same as estimating the degree to which the event is understood. In considering the estimation method, we consider the hypothesis remarkably accurate. Through discussions integrating the frameworks of deep learning and causal inference, it may be possible to realize an agent that can estimate the thoughts of others.

\subsection{Expression of explanation generation}
In this study, the important elements to be explained were extracted and presented to the user as they were. Because the purpose of an explanation is to bridge the gap between the concerned parties, it is not necessary to present information in a predetermined format. However, linguistic communication is often useful for humans. Examination of a method wherein important elements are converted into the form of language and presenting them to the user is a future task. On the other hand, beyond language, the modality and form of information presentation that might lead to a better explanation is an important issue to be further researched.

\section{Conclusion}
\label{chap:結論}
In this study, we proposed an MDP-based explanation generation framework to present the progress of an autonomous agent from the current state to the target state in a form that is easily accepted by humans. To generate an explanation that is easily acceptable to humans, it is important to present the minimum elements necessary to understand the approach to achieving the goal. Therefore, we proposed a method for identifying subgoals and estimating the user's understanding of events on a world model. In experiments where explanations were given to agents that had not learned a policy in a virtual environment, it was found that the subgoals derived using the proposed method were the minimum elements important for understanding the progress to the goal. In addition, subject experiments revealed that the generated explanation was a good summary of the approach to the goal, and that a high evaluation was obtained for the explanation of the reason for an action.

The most important issue in the future is how to realize a world model that assumes an MDP framework in solving real-world problems. The content and particle size of the generated explanations vary significantly, depending on the world model of the agent. Therefore, it is extremely important to find a method for agents to acquire a world model that is interpretable to humans. Furthermore, it is expected that it will be possible to generate explanations that are more acceptable to humans by improving the method of estimating the user's understanding of events.

\section*{Acknowledgement}
This work was supported by the New Energy and Industrial Technology Development Organization (NEDO).

\bibliographystyle{junsrt}
\bibliography{weekly.bib}

\end{document}